\documentclass{article}
\usepackage{arxiv}
\usepackage[flushleft]{threeparttable}
\usepackage[table]{xcolor}

\usepackage{times}
\usepackage{soul}
\usepackage{url}
\usepackage[hidelinks]{hyperref}
\usepackage[utf8]{inputenc}  
\usepackage[small]{caption}
\usepackage{graphicx}
\usepackage{amsmath}
\usepackage{amsthm}
\usepackage{booktabs}
\usepackage{algorithm}
\usepackage{array}
\usepackage{algorithmic}
\usepackage{multirow}
\usepackage{amssymb}
\usepackage[colorinlistoftodos]{todonotes}
\usepackage[switch]{lineno}

\usepackage{color}
\usepackage{subfig}
\usepackage{soul}

\newcolumntype{g}{>{\columncolor{lightgray!30}}}

\title{Computationally Assisted Quality Control for Public Health Data Streams}

\author{
Ananya Joshi\and
Kathryn Mazaitis\and
Roni Rosenfeld\and
Bryan Wilder
\affiliations
Carnegie Mellon University\\
\emails
\{aajoshi, kmazaitis, rrosenfeld, bwilder\}@andrew.cmu.edu
}

\begin{document}

\maketitle

\begin{abstract}
Irregularities in public health data streams (like COVID-19 Cases) hamper data-driven decision-making for public health stakeholders. A real-time, computer-generated list of the most important, outlying data points from thousands of daily-updated public health data streams could assist an expert reviewer in identifying these irregularities. However, existing outlier detection frameworks perform poorly on this task because they do not account for the data volume or for the statistical properties of public health streams. Accordingly, we developed FlaSH (\textbf{Fla}gging \textbf{S}treams in public \textbf{H}ealth), a practical outlier detection framework for public health data users that uses simple, scalable models to capture these statistical properties explicitly. In an experiment where human experts evaluate FlaSH and existing methods (including deep learning approaches), FlaSH scales to the data volume of this task, matches or exceeds these other methods in mean accuracy, and identifies the outlier points that users empirically rate as more helpful. Based on these results, \href{https://github.com/cmu-delphi/covidcast-indicators/tree/main/_delphi_utils_python/delphi_utils/flash_eval}{FlaSH} has been deployed on data streams used by public health stakeholders.
\end{abstract}

\section{Motivation and Introduction}
\begin{figure}%
    \centering
    \includegraphics[width=8.5cm, trim=0.5cm 2cm 0.5cm 2.5cm]{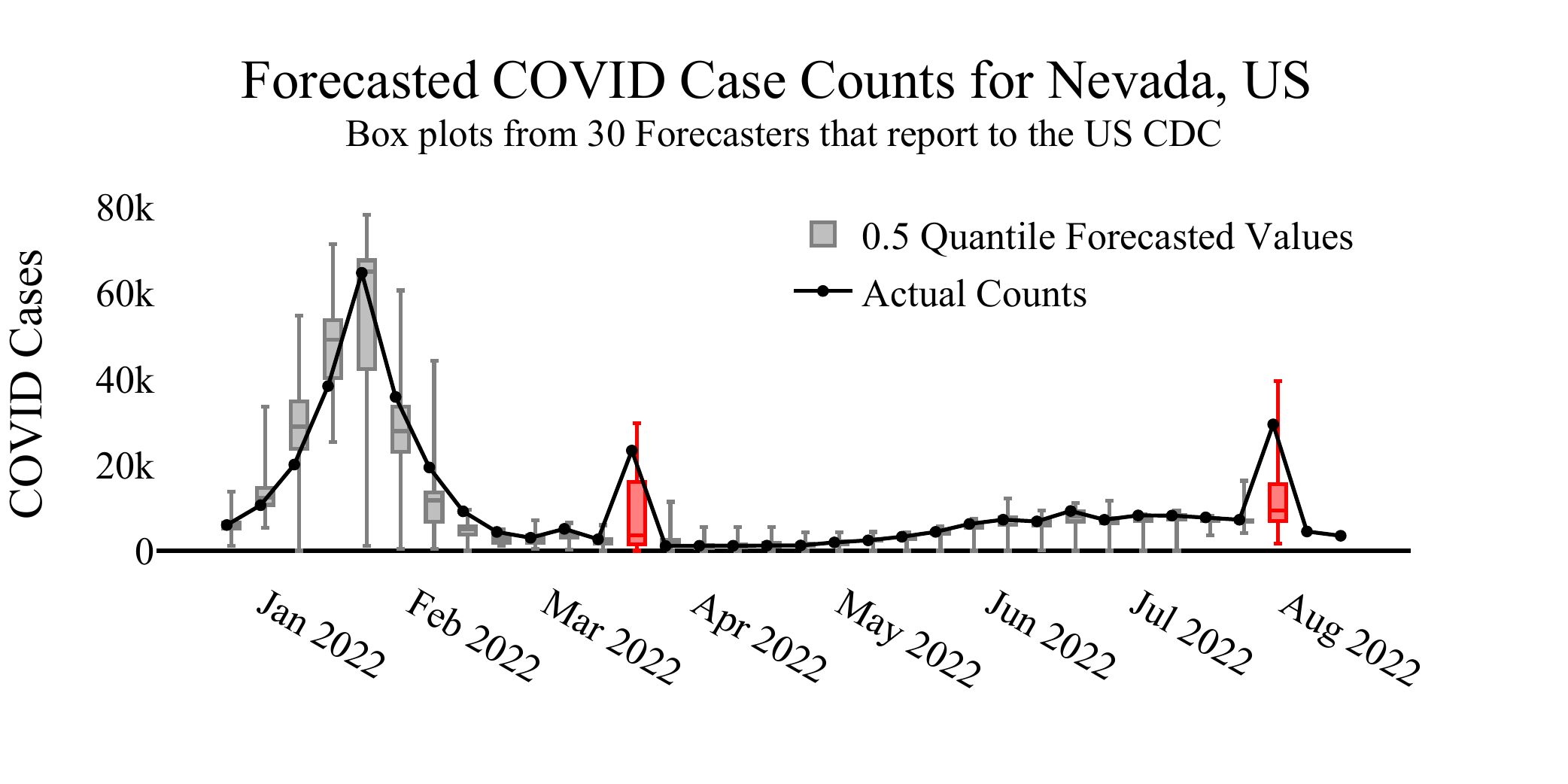} 
    \caption{Temporal irregularities in actual case counts, shown by the large spikes in March and July 2022, when cases were trending down, resulted in similar spikes for predicted counts (highlighted in red) that were then sent to the US Centers for Disease Control and Surveillance.}%
    \label{fig:nevada}%
\end{figure}

During the COVID-19 pandemic, daily-updated real-time public health data was used directly ~\cite{CDC_recs} or as input to methods that informed critical healthcare decisions and policies \cite{yu2021data} in support of Sustainable Development Goals such as good health and well being. However, aspects of public health data have hampered this data-driven decision-making in several ways. These include issues like data delays, corrections, and recording errors \cite{dong2022johns,saez2021potential} that may have masked important trends in disease progression \cite{kreps2020model}, as shown in Fig.\ \ref{fig:nevada}. Additionally, COVID variants or policy changes often cause sudden, notable distribution shifts in the data \cite{zhu2021revealing}. Finally, public health data streams are known to be biased or incomplete \cite{leslie2021does}. For example, regions with low healthcare resource availability may not have accurate COVID case counts. 

Addressing these issues is a significant challenge for any organization that curates public health data streams \cite{kraemer2021data}, including the Delphi Group at Carnegie Mellon University (Delphi). Delphi employs a team of full-time developers, statisticians, researchers, and product managers to maintain an accurate and performant public health data source\footnote{Delphi's open source repositories can be found at https://github.com/cmu-delphi}. Delphi's \href{https://cmu-delphi.github.io/delphi-epidata/api/covidcast.html}{publicly available API} \cite{farrow2015delphi} and other data products are regularly used by public health authorities in the United States (US), along with researchers, forecasters, journalists, and other users (totaling visits from over 78k unique IP addresses in January 2022). These stakeholders recommended that Delphi continuously monitor their data streams for irregularities so that Delphi's data users have more information about data quality issues, the state of the pandemic, and changes in regional disease behavior, to directly support data-driven decision-making.

To act on this recommendation, expert human reviewers in Delphi would need to regularly monitor at least ten thousand data streams for stakeholders (e.g.\ cases, deaths, and hospitalizations, at several geographical resolutions, including county, state, territory, and national level resolutions). If done manually, this type of monitoring is prohibitively expensive \cite{kraemer2021data}. Even if it were feasible, trained reviewers frequently miss critical irregularities due to the sheer reviewing load. While some outliers are so extreme that they require no human review, many outliers that signify irregularities are more nuanced and require close human attention. Computationally assisted quality control, where a reviewer only inspects the top entries from a computer-generated ranked list of outlier streams that a human should review, is promising because it could prioritize the reviewer's time for irregularity detection while retaining the trust and expertise a reviewer brings. 

Creating this computationally ranked list of outliers in public health streams is a difficult task. In addition to the practical constraints of operating over the large data volume necessitated by this task, outlier detection methods must be robust to the statistical \textit{noise}, \textit{nonstationarity},  \textit{day of week effects}, and \textit{limited historical data} that are prevalent in public health streams in order to provide helpful recommendations \cite{mcdonald2021can,reinhart2021open,wang2021comparing}. Further, the outlier detection methods must be simple and intuitive for reviewers to understand and trust them on this task. 

To address these challenges, we present FlaSH (\textbf{Fla}gging \textbf{S}treams in public \textbf{H}ealth), available open source at \url{https://github.com/cmu-delphi/covidcast-indicators/tree/main/_delphi_utils_python/delphi_utils/flash_eval}. FlaSH is a new outlier detection framework that produces a \textit{ranked list of recent values from data streams that most warrant human inspection}. FlaSH uses simple, scalable, and intuitive models to explicitly capture the statistical properties of public health data. To address challenges in evaluating unsupervised outlier detection methods in time series data like FlaSH, we also developed and conducted a classification and ranking evaluation of FlaSH's performance using input from several expert human reviewers. This is especially important given that many recent works in anomaly detection use semi-synthetic or simulation evaluations that may not truly reflect an expert user's assessment of the method utility. In this evaluation, FlaSH matches or outperforms previous outlier detection methods, including recent deep learning baselines.

\section{Practical Irregularity Detection Goals}
Our goal is to develop a framework that assists reviewers in detecting important irregularities in Delphi's data streams on behalf of public health data users. The data streams available to the Delphi Group vary by source (local governments, hospitals, private companies, and surveys), and each source has its own dynamics and measurement definitions. For example, the Johns Hopkins Centers for System Science and Engineering (JHU CSSE) COVID-19 source only curates data from publicly available reports \cite{dong2022johns}. They also only report real-time cumulative estimates. Thus, subsequent corrections to the cumulative figures can appear as large spikes or even negative values in derived daily case counts.

Detecting such irregularities across many sources is uniquely challenging for typical outlier detection methods, leading to a range of failure modes observed in our experiments. First, modern deep learning methods for outlier detection struggle with the large number of time series, each with a short history and rapid distribution shifts \cite{paleyes2022challenges}. To perform well, these highly parameterized models require long training histories often unavailable in public health settings. Moreover, high computational costs mean these methods scale poorly to real-time operation over thousands of distinct time series. Second, simpler statistical methods are not attuned to the specific structure of public health data and struggle to accurately identify irregularities \cite{wong2004data}. Third, neither class can leverage features of public health data streams that could assist with diagnosing irregularities. Because of these limitations, Delphi currently relies on volunteers and group members to \textit{manually} report issues on all data streams as they encounter them, but this process is unsystematic and expensive.

To start, the proposed outlier detection method should detect specific types of outliers present in public health streams that are relevant to Delphi's stakeholders so that the method is both context and user-dependent \cite{sejr2021explainable}. To identify these outlier categories, we conducted an exploratory analysis on data streams\footnote{The streams were from National, Texas, New York, LA County (CA), and Loving County (TX) sourced from JHU CSSE, Department of Health and Human Services, Google, and USA Facts.} of COVID Case Counts and Ratios, COVID Deaths, Hospital Visits, Google Symptoms Trends, and Doctors Visits at the national, state, territory, and county level resolutions from the first available date of the streams until December 2021. Using these streams, which each have a different possible range of values based on the region's population and the measurement quantity, we defined the following categories of outliers based on their ability to assist reviewers with identifying irregularities: 

\paragraph{Out of Range Values and Global Outliers.} These outliers are typically due to retrospective updates made by a data source in the value of a cumulative quantity. Out of range examples include ``negative" new cases (if the cumulative total was revised down) or more cases reported on a day than the population of the geographic area due to multi-day batched reports. Similarly, global outliers usually appear as large positive or negative spikes, but the `global' outlier thresholds may change over time as rapidly shifting disease dynamics undermine static thresholds. Still, both of these outliers are relatively easy to identify and very rarely require human review. 

\paragraph{Day of Week Outliers.} Many public health streams have systematic \textit{day of week effects} \cite{reinhart2021open}. For example, fewer COVID cases are reported on weekends partly because fewer people test on weekends. Day of week outliers occur when reported data points are anomalous relative to the expectation for their day of the week (even if they are within distribution for the stream as a whole). Unlike out of range or global outliers, day of week outliers are more difficult for humans to notice but may still indicate an irregularity in the stream. 

\paragraph{Trendline Outliers.} Data that deviate strongly from the recent trend (e.g.\, case counts were rising last week, but today's count is low) or from the recent trends of close geographic regions warrant attention. These phenomena are the most difficult for humans to detect and can indicate critical irregularities in the context of recent data.

To address failure modes from existing methods and detect these outlier categories, the proposed method must be intuitive, scale to the data volume, and provide outputs (outlier scores) that are correct and complement human judgment in this task. Practically, this requires the method to be a single-pass, point detection algorithm that integrates explainable AI and human computing interaction insights. Further, the ranking for the FlaSH list shown to the expert reviewers will be based on the trendline outlier scores because they can indicate critical irregularities in the context of recent data. Reviewers will also benefit from inspecting the global and day of week outlier scores reported alongside. Finally, each of these desired criteria must be evaluated to justifiably compare different outlier detection methods for this task.

\section{FlaSH Outlier Detection Method}

FlaSH formalizes the outlier detection problem discussed in the previous section as a model-based hypothesis test \cite{blazquez2021review}. We denote a single data stream as a time series $X_t$, $t = s...T$. Here, $s$ is the starting time for the stream analysis\footnote{Often, there is a ramp-up period before streams report reliable measurements, so we do not start at t=0.}, and $T$ is the current time. When it is necessary to discuss multiple geographic regions, we use $X^r$ to denote the stream for a given quantity in geographic region $r$ (e.g.\, the stream of COVID cases in a given US county). 

Suppose that $X_{s:T-1} \sim m$ for some $m \in \mathcal{M}$, where $\mathcal{M}$ is a set of models. We test the hypothesis that the most recent point in the stream is drawn from the same model ($H_0: X_T \sim m$). If the observed data has a low probability under this hypothesis, it means that $X_T$ was likely not generated from the same model $m$ as the historical data. This sudden shift from the data-generating distribution indicates a potential irregularity. We conduct the hypothesis test by first calculating a test statistic measuring the discrepancy between observed values and values predicted by $m$. We then obtain a $p$-value by comparing the real-time test statistic value to a historical distribution of test statistics $\mathcal{P}$. FlaSH instantiates this entire method via a sequence of 3 steps: 

\paragraph{S1: Process Data.} We want to fit a model $m$ such that points with irregularities appear in the most extreme tails of the $m$'s predictive distribution. However, training $m$ on out of range, global, and day of week outliers both distorts the model and inflates the tails of the distribution of prediction error so that more subtle deviations no longer stand out. We process the data to identify and impute these outliers prior to training. The key challenge in this step is to accommodate the statistical properties of public health data.

\paragraph{S2: Obtain Predicted Values.} After processing, we fit a parametric model $m$ from a model class $\mathcal{M}$ that uses the history of the stream to predict future values. Choosing an appropriate $\mathcal{M}$ is nontrivial. Heavily parameterized models, like many deep learning models, are unsuitable because of the limited data history available to tune the model, the expensive ground truth labels, and the rapid distribution shifts in the types of irregularities per stream. Further, stakeholders prioritize interpretability, so the model class must be intuitive.

\paragraph{S3: Compare Predicted and Observed Values.} Finally, FlaSH compares the observed and predicted values to test if $X_T$ could have been generated from $m$ given the historical performance of observed and predicted values. The critical decision in this step is the choice of the test statistic and construction of its distribution under the null hypothesis, which are complicated by short training histories and the resulting need to share information across geographic regions.

\begin{figure}%
    \centering
     \includegraphics[width=8cm , trim=0.5cm 1.4cm 0cm 0cm, clip]{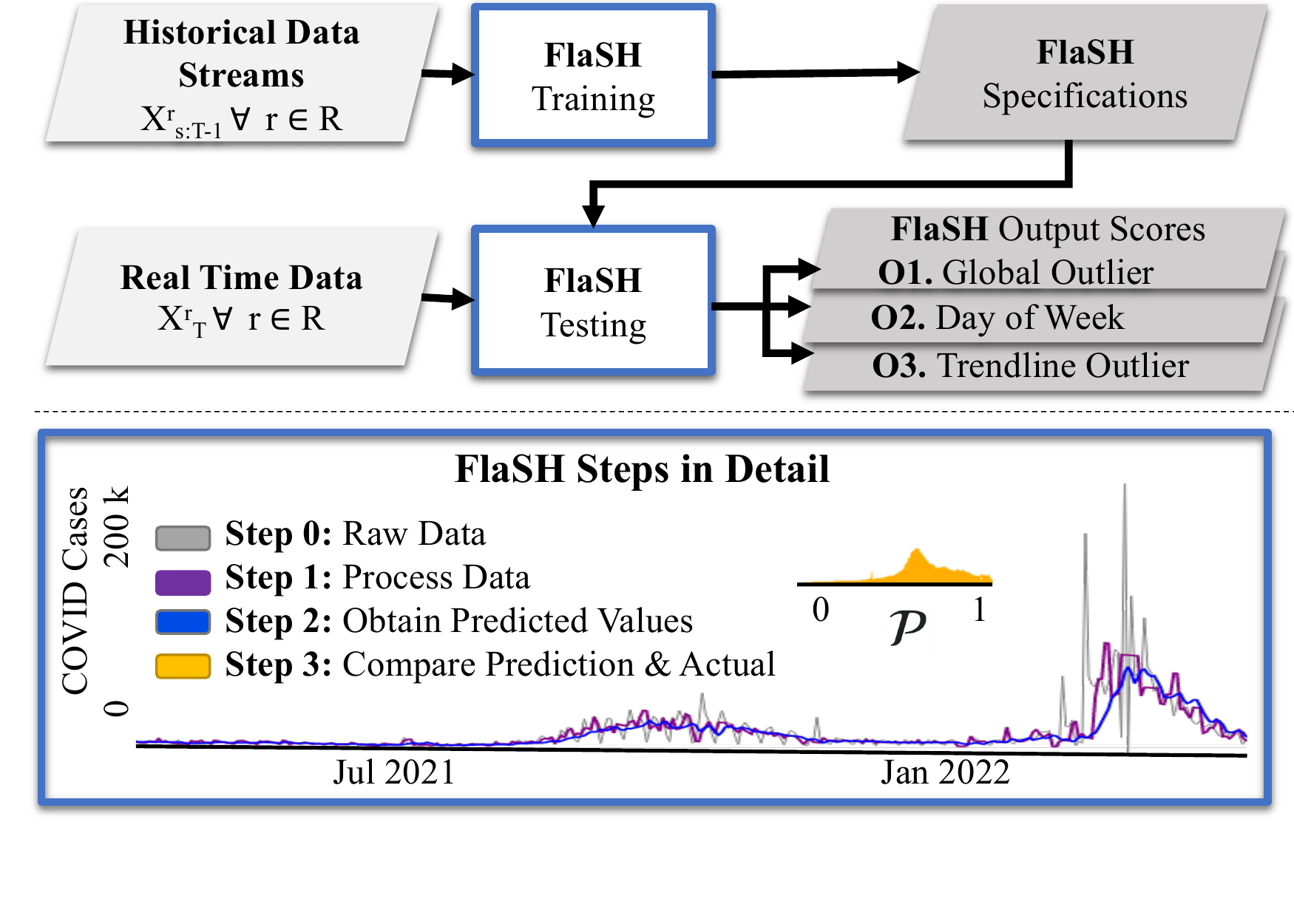} 
    \caption{In the FlaSH outlier detection method, data stream inputs are processed through FlaSH to generate informational outlier scores. FlaSH itself has three steps. The raw data (gray) is processed [S1] (purple), and model $m$ is used to predict future values [S2] (blue). Then, the historical performance of model $m$ is captured with the test statistic distribution (gold), and this distribution is used to compare predicted and actual values [S3].}%
    \label{fig:diagram}%

\end{figure}

We now discuss each step, as displayed in Fig.\ \ref{fig:diagram}.

\subsection{Process Data}
Trendline outliers cannot be reliably identified if the model is trained on data that also includes out of range, global, and day of week outliers. However the thresholds for determining these outliers change with distribution shifts in the stream. To address this challenge, first, different COVID regimes, or waves, are identified via changepoints. Then, within each regime, existing outliers are detected and imputed. 

\subsubsection{Identifying Changepoints in Nonstationary Streams}
Values that would be outliers when there is no COVID wave may not be outliers during a COVID wave. This phenomenon of distinct, underlying waves, or regimes, in public health streams is why they are known to be statistically \textit{nonstationary} \cite{chimmula2020time}\footnote{Operationally, we consider regimes present in streams that are updated daily with at least 60 historical data points. On streams with fewer than 60 data points, we provide interquartile range-defined outliers.}. To identify these regimes in historical data, FlaSH uses the Pelt Changepoint Algorithm \cite{killick2012optimal,ruptures}, parametrized with a Gaussian model and a minimum of four weeks between change points\footnote{Four weeks is the maximum horizon for many short-term forecasts \cite{cramer2022united}, likely because health dynamics change drastically after that horizon.}. However, individual streams may be very noisy, and Pelt sometimes overfits to this noise to return regimes inconsistent with expert knowledge of disease dynamics. Therefore, we take advantage of geographical dependencies\footnote{Data reporting and health policies are generally consistent at the state level \cite{simon_2021}.} by searching for changepoints that are jointly applicable across a set of nearby regions. Specifically, we run the Pelt algorithm on the streams for all counties within a given state, jointly optimizing Pelt's objective across these regions to find changepoint days that describe the regimes well across these streams. 

While Pelt can identify changepoints in historical data, it does not identify if real-time data represents a changepoint. Instead of retraining FlaSH daily to find new changepoints, which would be computationally expensive, FlaSH assumes there is no changepoint until there is sufficient evidence of nonstationarity to trigger retraining as follows. Under the null hypothesis, there is no changepoint, and $p$-values are uniformly distributed by definition. If the distribution of the test statistic for the hypothesis test $H_0$ significantly shifts, then the Kolmogrov-Smirnov test can identify whether the empirical p-value distribution since the last retraining deviates significantly from the uniform distribution. The user can select the test significance level $\alpha$ according to their desired trade-off between the computational expense of retraining and increased accuracy. Even if a new changepoint is not detected, FlaSH is retrained every 3 months, which roughly corresponds to a change in season, and the expert reviewer can retrain at any time. 

\subsubsection{Identifying Outliers Within Regimes}
Within each changepoint-defined regime, FlaSH identifies out of range, day of week, and global outliers, and it imputes non-outlying values that are later used for modeling. First, out of range outliers, like negative COVID Cases, are identified and imputed to be in range. Second, data is separated by day of week, and points where $\lvert z_{score} \rvert \geq 3$ with respect to the points for that day of the week in the regime are identified as day of week outliers. The imputed value for downstream analysis is the median value of the difference in day of the week added to the value for the prior day. Day of week sensitivity is important here because of systematic patterns across the week, like that the median value for Sundays is usually lower than the median for Tuesdays. 

Third, we process the time series to remove systematic day of week effects (unlike the previous step, which handled points far outside the typical pattern for their weekday). FlaSH uses a Poisson regression method $w$ (part of Delphi's public API\footnote{\url{https://github.com/cmu-delphi/covidcast-indicators/blob/main/_delphi_utils_python/delphi_utils/weekday.py}}) which outputs a weekday-corrected value $w(X_t)$. This model removes systematic differences in mean values across days (e.g.\, by scaling values on Saturday up and scaling Mondays down) to obtain a time series without day of week effects. Removing such systematic periodicity enables downstream predictive models to fit the data-generating process using fewer parameters. 

Finally, after the day of week correction, FlaSH addresses the \textit{noisiness} of the stream by identifying global outliers in the day of week corrected data as those with $\lvert z_{score} \rvert \geq 3$, calculated from all weekdays in the day of week corrected data. These points are imputed using the mean value of the current regime. 
Having removed out of range, global, and day of week outliers, FlaSH treats the processed data across all regimes as the null distribution and can now identify trendline outliers as specified by the following two steps of FlaSH. 

\subsection{Obtain Predicted Values}

To identify trendline outliers, FlaSH uses a small sample of the processed historical data to train a predictive model $m$ for $X_T$ from model class $\mathcal{M}$ and then uses the remaining processed historical data to characterize the performance of the model. Specifically, the training set for FlaSH's null hypothesis model is the maximum of 10\% of the historical data or 30 points. FlaSH then uses $\mathcal{M}$ : Linear Autoregressive (AR) models (lag=7) , where $m$ is characterized by the linear weights, $\hat{\beta}$, fit during training. This class of models is preferred in public health applications for its simplicity and performance with \textit{limited historical data} \cite{mcdonald2021can}. The remaining processed historical data (not used to fit the model) is used to generate predictions $\hat{X_t}$. 

\subsection{Compare Predicted and Observed Values}

Models from any model class $\mathcal{M}$ fit with the null historical data will not perform uniformly across all streams. Accordingly, out-of-sample data is essential to quantify the typical discrepancy between model predictions and observed values per stream. Outliers can then be identified when the discrepancy between predictions and observations is more than typical, as determined by a distribution of historical performance. For example, if a model consistently predicts higher values than what is observed, then the outlier score should reflect the fact that $\hat{X}_T > X_T$ is not surprising.

To quantify the discrepancy between predicted and observed values, let $N^r$ denote the total population of geographic region $r$. The day of week corrected observed values ($w(X^{r}_{t})$, corrected to be comparable to the predicted values) and the predicted values ($\hat{X^{r}_{t}} = \hat{\beta}* w(X^{r}_{t-1:t-7})$) are used to calculate the test statistic $k_{t}$:
\begin{align*}
    k_{t} = (P(w(X^{r}_t) < D))\\
    D \sim \text{Bin}\left(n=N^r, p=\frac{\hat{X^{r}_{t}}}{N^r}\right)
\end{align*} 
This test models the counts in a region as a binomial distribution $D$. The probability of infection per person is the number of predicted counts divided by the region's population size. Intuitively, we test the hypothesis that the actual observed counts are drawn from a distribution parameterized by our predictions. Extreme values of the test statistic indicate that the observations were much bigger or smaller than expected given the predictions. Each stream model's typical performance discrepancy is specified by a distribution $\mathcal{P}^r$, composed of test statistics $k^{r}_{30:T-1}$, that compares observed values and the predicted values for the out-of-sample historical data $X^{r}_{30:T-1}$. However, there is often too little history to approximate the null distribution of an individual stream effectively, with a minimum of 30 points characterizing each distribution if there are only 60 days of historical data. Accordingly, we define the pooled test statistic distribution $\mathcal{P}$, specified by $\bigcup_{r \in R}k^{r}_{30:T-1}$, where $R$ is all the counties in a state if $r$ is a county, else $R$ is all states and territories in a nation, because these streams share geographic context. Note that pooling is enabled by the design of our test statistic, which is chosen to ensure comparable distributions across regions (e.g.\ via normalizing by the population). 

\subsection{FlaSH Output} The final output of FlaSH is a list of real-time points ranked by how extreme their test statistic is via the transformation $|2p-1|$, where $p$ is the $p$-value for the real-time test statistic in the pooled historical test statistic distribution \textbf{$\mathcal{P}$}. This transformation ensures that the most outlying points (from either distribution tail) will top the ranked list. 

\section{FlaSH Labels, Evaluation, \& Feedback}

 As noted in the literature, accurately evaluating algorithms for unsupervised time series outlier detection is challenging \cite{wu2021current}. In most previous work, human-generated labels have not been provided by experts (instead coming from readily available subjects such as students or Mechanical Turk workers). Non-expert labels are noisy since identifying outliers often requires domain-specific knowledge. However, outlier detection method performance on simulated data or data with synthetically injected outliers \cite{lai2021revisiting} rarely translates to practical performance on real-world data in epidemiology generally \cite{wong2004data}.
 
 One of our key contributions is to address this limitation in the outlier detection literature via a rigorous, real-world evaluation of outlier detection methods for public health data. We obtained high-quality labeled data from human subject matter experts- Delphi members who are directly involved in building statistical or software systems using public health data and who regularly encounter the impact of data irregularities. In contrast to the binary labels standard in previous work, which may not be sufficient as different experts have different thresholds for outlier determination, asking experts to rank outliers that warrant human inspection provides a more informative comparison for FlaSH's output.

 For additional evaluation rigor, we preregistered the FlaSH version, survey design, and analysis before data collection began \cite{OSF}. This ensures that our algorithm was finalized before any data collection occurred, giving an unbiased (prospective) evaluation of FlaSH's performance. Real-time COVID Case data streams (3341 streams at county, state, territory, and national levels available from May 2020-May 2022) were initially sourced daily from JHU CSSE. Of these, five streams, including the national stream, were randomly chosen from the following sets to ensure stream variety for the evaluation: the top 10\% of populous states (Pennsylvania), bottom 90\% of populous states (Arkansas), top 10\% populous FIPS regions (36081 Queens County, NY), and bottom 90\% populous FIPS regions (72043 Coamo Municipality, PR), as per the US Census. 

\begin{figure}
\centering
\includegraphics[width=9cm, trim=0cm 1.4cm 0cm 0.5cm, clip]{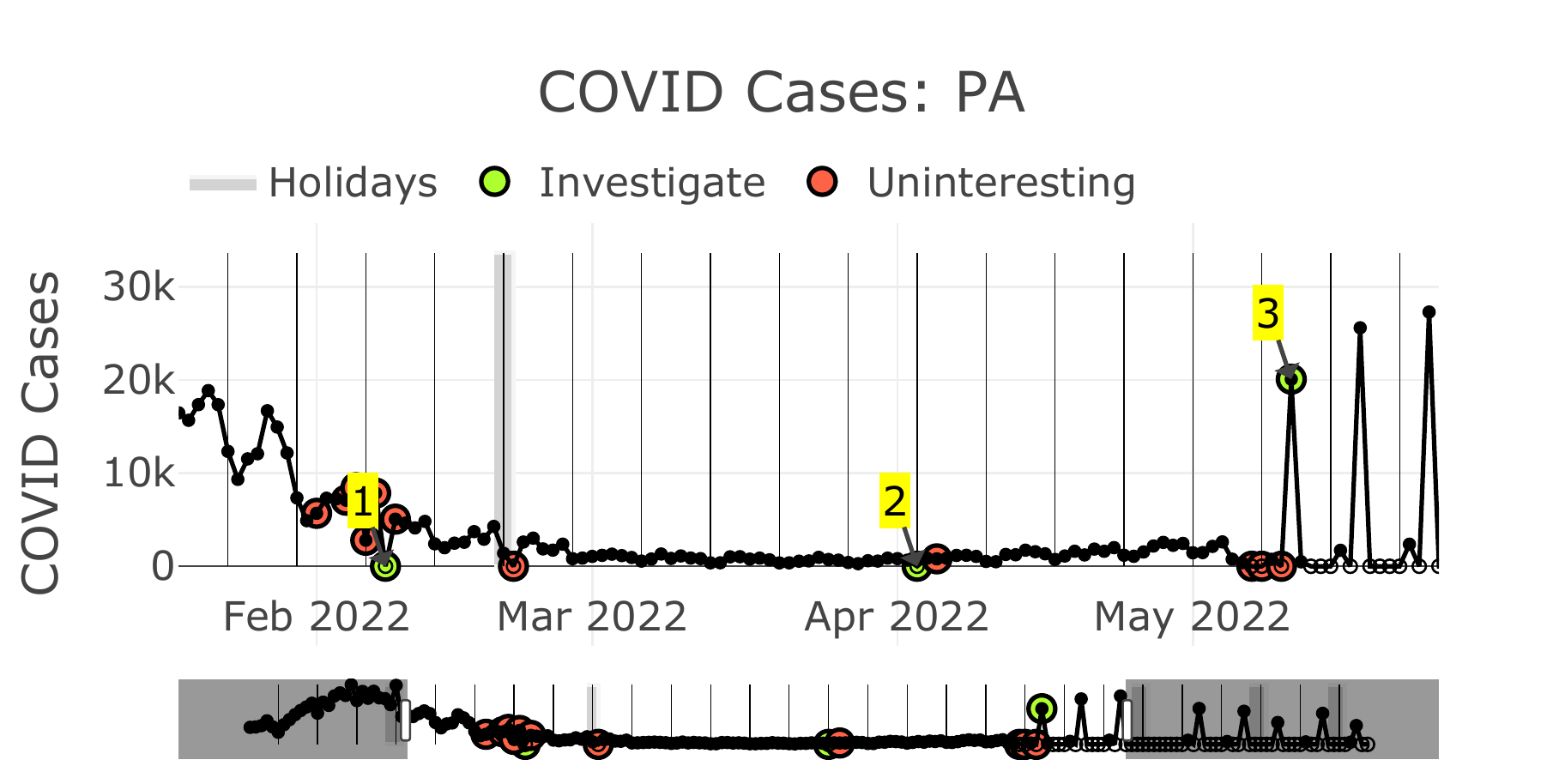}
\caption{Example of a Survey Task. Respondents click on the time series plot to mark points as unevaluated, uninteresting, or warrants investigation. They also rank points that warrant investigation, and these rankings appear on the plot in yellow. Respondents could zoom, pan, and see a 7 day average per graph. }
\label{fig:survey}
\centering
\end{figure}

\subsection{Survey and Analysis} To gather ground truth in order to understand FlaSH's performance on empirical data, we designed a \href{https://github.com/Ananya-Joshi/IJCAI23_Supplemental}{web survey} that has 10 interactive questions (Fig.\ \ref{fig:survey}) \footnote{\url{https://github.com/Ananya-Joshi/IJCAI23_Supplemental}}. First, respondents classify candidate data points from a public health stream as `warrants human investigation' or `uninteresting'. Then, they are asked to rank (with possible ties) the subset of these candidates they think would warrant additional human inspection. 

To form this candidate set of evaluation points, we needed to select stream values that are at least somewhat anomalous. That way, survey respondents could meaningfully \textit{distinguish} between points that are potentially anomalous. In practice, we expect $<$ 1\% of all points in a stream to represent irregularities. Accordingly, this candidate set is formed by taking the union of the top outlying points output by both FlaSH and 8 previously proposed outlier detection methods given all historical data (see Sec. \ref{results}). By filtering to data points at the top of at least one algorithm, the candidate set is limited to points that are considered anomalous by some method. This empirically meant the candidate set comprised of points that were at least interesting enough to classify and rank.

In Questions 1-5 (Q1-5), the candidate set was formed from the top 5\% of points from at least one algorithm for each of the 5 possible data streams. In Q6-10, respondents were asked to reconsider the top 2\% of points from at least one algorithm, a subset of these candidates from Q1-5, in more detail to test for respondent internal consistency. They were also asked how likely they would have flagged each point for human review had it not been identified by an algorithm (`unlikely', `somewhat unlikely', `neither', `somewhat likely', or `likely'). This allows us to measure the value added by the algorithm over what would have been obvious to a human. 

We evaluate the algorithm's performances in a realistic setting of only 60 days of history for training (12/21/2021-1/31/2022). Our test set was the following 100 days (2/1-5/12/2022). To compare the survey results to the outlier detection method outputs, we use a range of metrics to capture the complexity of real-world outlier detection. Both traditional binary classification and ranking metrics provide information on how well the system finds points that the majority of respondents thinks warrants human inspection. We also seek to understand which points from outlier detection algorithms provided the most benefit to users based on self-reports. The points which were rated as both highly anomalous and unlikely to have been flagged without an algorithm are the most valuable potential contributions of computationally assisted quality control. 

\paragraph{Survey Quality.} The total number of survey participants (n=13) is a significant increase over previous work (e.g.\ n=3 in \cite{wong2004data}). We tested for internal consistency in respondents that answered both sets of survey questions (Q1-5 and Q6-10) by measuring response centrality between the Copeland aggregate per paired question (e.g. Q1 \& Q6, Q2 \& Q7) and the raw ranks per person. High centrality values (0.83 $\pm$ 0.13) suggest that respondents generally were consistent in their pairwise preferences between the two sets. Still, the average number of points that warranted inspection per person varied from $<2$ to $>6$, supporting that the \textit{threshold} for identifying points of interest varies greatly by individual and reinforcing the importance of sampling a wider range of experts than historical standards suggest.

\section{Results and Analysis}\label{results} 

\begin{table*}
    \centering
    \begin{threeparttable}
    \begin{tabular}{>{\centering\arraybackslash}p{0.2cm}p{0.1cm}|p{0.85cm}p{1.15cm}g p{1.5cm}|p{1.1cm}p{1.1cm}p{0.9cm}p{0.9cm}p{1cm}p{0.9cm}p{1.2cm}}
        \toprule
        \multicolumn{2}{c}{Model Class} & \multicolumn{3}{c}{\textbf{AR}}  &  \textbf{DeepLog} & \textbf{Telem.} & \textbf{VAE} & \textbf{LOF} & \textbf{LODA} & \textbf{IF} & \textbf{KNN} \\
        \multicolumn{2}{c}{Implementation} & {TODS} & {Mixed$\dagger$} & \multicolumn{1}{c}{\textbf{\ul{FlaSH}}} & \multicolumn{6}{c}{TODS} \\
        \midrule
        \hline
        \multicolumn{2}{c|}{Training (s)}  & \multicolumn{2}{c}{\strut{$10.1$}\hbox{\strut\fontsize{4pt}{4pt}$\pm0.3$}} &{\strut{$169$}\hbox{\strut\fontsize{4pt}{4pt}$\pm0.8$}}
         & DNF & DNF & DNF & {\strut{$8$}\hbox{\strut\fontsize{4pt}{4pt}$\pm0.2$}} & {\strut{$71$}\hbox{\strut\fontsize{4pt}{4pt}$\pm0.1$}}
   & DNF & {\strut{$7$}\hbox{\strut\fontsize{4pt}{4pt}$\pm0.08$}} \checkmark\\ 
        \midrule
        \midrule
        
        
        \multirow{4}{*}{\rotatebox[origin=c]{90}{\parbox[c]{1.1cm}{\centering Binary}}}&
        Accuracy& {\strut{$0.78$}\hbox{\strut\fontsize{4pt}{4pt}$\pm0.02$}} & {\strut{$0.71$}\hbox{\strut\fontsize{4pt}{4pt}$\pm0.04$}} &
        {\strut{$0.8$}\hbox{\strut\fontsize{4pt}{4pt}$\pm0.03$}} \checkmark&{\strut{$0.8$}\hbox{\strut\fontsize{4pt}{4pt}$\pm0.04$}}\checkmark&{\strut{$0.6$}\hbox{\strut\fontsize{4pt}{4pt}$\pm0.04$}}&{\strut{$0.76$}\hbox{\strut\fontsize{4pt}{4pt}$\pm0.04$}}&{\strut{$0.69$}\hbox{\strut\fontsize{4pt}{4pt}$\pm0.01$}}&{\strut{$0.68$}\hbox{\strut\fontsize{4pt}{4pt}$\pm0.04$}}&{\strut{$0.79$}\hbox{\strut\fontsize{4pt}{4pt}$\pm0.04$}}&{\strut{$0.74$}\hbox{\strut\fontsize{4pt}{4pt}$\pm0.03$}}\\
        
{}&Bal.Acc.&{\strut{$0.68$}\hbox{\strut\fontsize{4pt}{4pt}$\pm0.02$}}&{\strut{$0.59$}\hbox{\strut\fontsize{4pt}{4pt}$\pm0.06$}}&{\strut{$0.73$}\hbox{\strut\fontsize{4pt}{4pt}$\pm0.05$}}\checkmark&{\strut{$0.72$}\hbox{\strut\fontsize{4pt}{4pt}$\pm0.05$}}&{\strut{$0.42$}\hbox{\strut\fontsize{4pt}{4pt}$\pm0.03$}}&{\strut{$0.67$}\hbox{\strut\fontsize{4pt}{4pt}$\pm0.07$}}&{\strut{$0.55$}\hbox{\strut\fontsize{4pt}{4pt}$\pm0.03$}}&{\strut{$0.54$}\hbox{\strut\fontsize{4pt}{4pt}$\pm0.05$}}&{\strut{$0.7$}\hbox{\strut\fontsize{4pt}{4pt}$\pm0.07$}}&{\strut{$0.62$}\hbox{\strut\fontsize{4pt}{4pt}$\pm0.05$}}\\

{}&F1&{\strut{$0.54$}\hbox{\strut\fontsize{4pt}{4pt}$\pm0.05$}}&{\strut{$0.43$}\hbox{\strut\fontsize{4pt}{4pt}$\pm0.09$}}&{\strut{$0.64$}\hbox{\strut\fontsize{4pt}{4pt}$\pm0.08$}}\checkmark&{\strut{$0.63$}\hbox{\strut\fontsize{4pt}{4pt}$\pm0.07$}}&{\strut{$0.19$}\hbox{\strut\fontsize{4pt}{4pt}$\pm0.07$}}&{\strut{$0.53$}\hbox{\strut\fontsize{4pt}{4pt}$\pm0.12$}}&{\strut{$0.33$}\hbox{\strut\fontsize{4pt}{4pt}$\pm0.08$}}&{\strut{$0.34$}\hbox{\strut\fontsize{4pt}{4pt}$\pm0.09$}}&{\strut{$0.56$}\hbox{\strut\fontsize{4pt}{4pt}$\pm0.11$}}&{\strut{$0.42$}\hbox{\strut\fontsize{4pt}{4pt}$\pm0.09$}}\\

{}&ROCAUC&{\strut{$0.79$}\hbox{\strut\fontsize{4pt}{4pt}$\pm0.02$}}& {\strut{$0.73$}\hbox{\strut\fontsize{4pt}{4pt}$\pm0.06$}} & {\strut{$0.75$}\hbox{\strut\fontsize{4pt}{4pt}$\pm0.06$}}&{\strut{$0.82$}\hbox{\strut\fontsize{4pt}{4pt}$\pm0.05$}}\checkmark&{\strut{$0.42$}\hbox{\strut\fontsize{4pt}{4pt}$\pm0.07$}}&{\strut{$0.68$}\hbox{\strut\fontsize{4pt}{4pt}$\pm0.06$}}&{\strut{$0.62$}\hbox{\strut\fontsize{4pt}{4pt}$\pm0.04$}}&{\strut{$0.44$}\hbox{\strut\fontsize{4pt}{4pt}$\pm0.07$}}&{\strut{$0.66$}\hbox{\strut\fontsize{4pt}{4pt}$\pm0.08$}}&{\strut{$0.65$}\hbox{\strut\fontsize{4pt}{4pt}$\pm0.07$}}\\

         \midrule
         \midrule
         \multirow{3}{*}{\rotatebox[origin=c]{90}{\parbox[c]{1.1cm}{\centering Ranking}}} &Distance &{\strut{$0.66$}\hbox{\strut\fontsize{4pt}{4pt}$\pm0.39$}}&{\strut{$1$}\hbox{\strut\fontsize{4pt}{4pt}$\pm0$}}&{\strut{$0.62$}\hbox{\strut\fontsize{4pt}{4pt}$\pm0.39$}}\checkmark&{\strut{$0.63$}\hbox{\strut\fontsize{4pt}{4pt}$\pm0.36$}}&{\strut{$0.83$}\hbox{\strut\fontsize{4pt}{4pt}$\pm0.24$}}&{\strut{$0.66$}\hbox{\strut\fontsize{4pt}{4pt}$\pm0.37$}}&{\strut{$0.66$}\hbox{\strut\fontsize{4pt}{4pt}$\pm0.39$}}&{\strut{$0.71$}\hbox{\strut\fontsize{4pt}{4pt}$\pm0.39$}}&{\strut{$0.67$}\hbox{\strut\fontsize{4pt}{4pt}$\pm0.39$}}&{\strut{$0.66$}\hbox{\strut\fontsize{4pt}{4pt}$\pm0.39$}}\\
{}&RBO&{\strut{$0.84$}\hbox{\strut\fontsize{4pt}{4pt}$\pm0.1$}}&{\strut{$0.89$}\hbox{\strut\fontsize{4pt}{4pt}$\pm0.08$}}&{\strut{$0.84$}\hbox{\strut\fontsize{4pt}{4pt}$\pm0.1$}}&{\strut{$0.84$}\hbox{\strut\fontsize{4pt}{4pt}$\pm0.1$}}&{\strut{$0.84$}\hbox{\strut\fontsize{4pt}{4pt}$\pm0.1$}}&{\strut{$0.89$}\hbox{\strut\fontsize{4pt}{4pt}$\pm0.07$}}&{\strut{$0.88$}\hbox{\strut\fontsize{4pt}{4pt}$\pm0.08$}}&{\strut{$0.93$}\hbox{\strut\fontsize{4pt}{4pt}$\pm0.06$}}\checkmark&{\strut{$0.91$}\hbox{\strut\fontsize{4pt}{4pt}$\pm0.11$}}&{\strut{$0.88$}\hbox{\strut\fontsize{4pt}{4pt}$\pm0.08$}}\\
        {}&Corr.&{\strut{$0.2$}\hbox{\strut\fontsize{4pt}{4pt}$\pm0.63$}}&{\strut{$0.42$}\hbox{\strut\fontsize{4pt}{4pt}$\pm0.45$}}&{\strut{$0.37$}\hbox{\strut\fontsize{4pt}{4pt}$\pm0.57$}}&{\strut{$0.43$}\hbox{\strut\fontsize{4pt}{4pt}$\pm0.54$}}\checkmark&{\strut{$-0.13$}\hbox{\strut\fontsize{4pt}{4pt}$\pm0.71$}}&{\strut{$0.18$}\hbox{\strut\fontsize{4pt}{4pt}$\pm0.64$}}&{\strut{$0.21$}\hbox{\strut\fontsize{4pt}{4pt}$\pm0.67$}}&{\strut{$0.24$}\hbox{\strut\fontsize{4pt}{4pt}$\pm0.69$}}&{\strut{$0.17$}\hbox{\strut\fontsize{4pt}{4pt}$\pm0.68$}}&{\strut{$0.22$}\hbox{\strut\fontsize{4pt}{4pt}$\pm0.66$}}\\
         \midrule
         \midrule
         \multicolumn{2}{c|}{Assistive Rank\tnote{*}} & 
         {\strut{$8.00$}\hbox{\strut\fontsize{4pt}{4pt}$\pm6$}} & {\strut{$3.66$}\hbox{\strut\fontsize{4pt}{4pt}$\pm1$}} &
         {\strut{$1.33$}\hbox{\strut\fontsize{4pt}{4pt}$\pm0.7$}} \checkmark&
         {\strut{$2.33$}\hbox{\strut\fontsize{4pt}{4pt}$\pm0.7$}}&
         {\strut{$41.33$}\hbox{\strut\fontsize{4pt}{4pt}$\pm38$}}&
         {\strut{$32.00$}\hbox{\strut\fontsize{4pt}{4pt}$\pm57$}}&
         {\strut{$24.00$}\hbox{\strut\fontsize{4pt}{4pt}$\pm40$}}&
         {\strut{$70.67$}\hbox{\strut\fontsize{4pt}{4pt}$\pm51$}}&
         {\strut{$47.33$}\hbox{\strut\fontsize{4pt}{4pt}$\pm39$}}&
         {\strut{$5.33$}\hbox{\strut\fontsize{4pt}{4pt}$\pm5$}}\\
        \bottomrule        
    \end{tabular}
    \begin{tablenotes}
\item[*] Mean rank of points somewhat unlikely or unlikely to be caught by human
\item[$\dagger$] Mixed model with FlaSH data processing [S1] and TODS comparison of predicted and observed values [S3].
\end{tablenotes}
    \renewcommand\thetable{5}
    \caption{Summary of Algorithm Comparison with 60 Days Historical Data. \protect\checkmark marks the best algorithm in each row.}
    \label{tab:algs}
\end{threeparttable}

\end{table*}

We compare the \textit{trendline outlier scores} from FlaSH to outlier scores from the following off-the-shelf outlier detection algorithm baselines implemented in TODS\footnote{Each algorithm had a setting 7 day windows where applicable to account for \textit{day of week effects}.} that span recent deep learning methods, classical machine learning, and statistical approaches: DeepLog \cite{du2017deeplog}, Telemanom (Telem.) \cite{telemanom}, Variational Autoencoder (VAE) \cite{an2015variational}, Local Outlier Factor (LOF) \cite{breunig2000lof}, Lightweight Online Detector of Anomalies (LODA) \cite{pevny2016loda}, Isolation Forest (IF) \cite{liu2008isolation}, k-Nearest Neighbors (KNN) \cite{angiulli2002fast}, and Linear AR Model \cite{gupta2013outlier}. These methods have in-built data processing [S1] and prediction comparison [S3] steps, just like FlaSH. We use default hyperparameters for the TODS implementations because there is not enough recent data to select hyperparameters. In fact, many of these models are too costly to even train once on the full set of streams, much less to do hyperparameter selection with many repeated runs. One strength of FlaSH is that it has no hyperparameters because it is designed for this task. 

Additionally, for an ablation study, we compare results from the TODS AR model implementation, which has the same model class $\mathcal{M}$ as FlaSH, to a mixed implementation (Mixed), where the processing step [S1] is the same as FlaSH, and the prediction comparison step [S3] is from TODS. 

\paragraph{FlaSH is computationally scalable.} We find that FlaSH easily scales to a large number of data streams, while many deep learning methods become infeasible. Performance statistics (Table \ref{tab:algs}) were reported from experiments using a 2.6 GHz 6-Core Intel Core i7 machine. Each algorithm was trained on the full 3341 JHU CSSE COVID-19 case streams with 60 days of history. This setup mimics the setting that we expect algorithms to scale to in deployment. A few algorithms (mainly deep learning algorithms) did not finish training within one day (DNF). Training time can only increase for these deep learning implementations as historical data increases. While GPU acceleration may benefit deep learning models, such specialty hardware may not be available in many public health settings. 

\paragraph{FlaSH performs well on outlier detection metrics.} Although many of the existing outlier detection methods have infeasibly long training times for daily deployment, we compare the performance of all algorithms using the labeled data from the survey. Table \ref{tab:algs} shows the 95\% CI of various traditional binary and ranking outlier detection metrics across all participants for Q1-5 per algorithm. 

In the binary analysis, points identified by the majority of respondents as to-investigate were marked as outliers (ground truth)\footnote{The base rates were: US (2/14), Pennsylvania (9/14), Arkansas (3/16), FIPS 36081 (6/24) and FIPS 72043 (5/21).}. To calculate binary labels from each algorithm to compare to this ground truth, we used the following process. Let $k$ denote the number of human-identified outliers for a stream. For each algorithm, we took the top $k$ points, ranked according to the algorithm's outlier scores, as the predicted outliers for binary classification tasks and compared these results to the ground truth labels. We report the 95\% CI metrics per person and per question for accuracy, balanced accuracy score, F1 score, and the ROC-AUC score. On average, FlaSH meets or exceeds the performance of all baselines in the binary analysis. FlaSH performs slightly better than DeepLog, an unusable, but performant, deep learning method. Some model classes like Telemanom and LODA performed poorly on the ROC-AUC score because while they identified global outliers very clearly, they failed to capture other kinds of outliers (e.g.\, trendline or day of week outliers). For the ranking analysis, each algorithm's ranking of the subset points available in Q1-5 that a majority of participants marked as warrants suspicion was compared to each respondent's rankings using Hamming distance (lower is better), Ranked-Biased Overlap (RBO) \cite{webber2010similarity}, and swap correlation (corr). Once again, FlaSH performs comparably to DeepLog and is competitive with the other algorithms. 

Finally, FlaSH shows strong improvements over the TODS AR implementation. While the TODS AR method is uncompetitive with other approaches, by using data processed using FlaSH's first step (Mixed), the AR model can better build a null model of the data. Still, because the TODS outlier scoring uses the absolute difference between the predicted and observed values to rank points, the mixed approach performs poorly on streams with small case counts, as reflected in the results. Compared to the TODS implementation with the same model class $\mathcal{M}$, FlaSH's processing [S1] and comparison [S3] steps together provide clear performance benefits. 

\paragraph{FlaSH can complement human judgment.} We find that FlaSH identifies useful points that were unlikely to have been inspected without computational assistance (via an algorithm identifying the point), as shown in the Assistive Rank section of Table \ref{tab:algs}. Specifically, we examine the set of points that (a) the majority of humans rated as warranting investigation after a full examination, and (b) at least 40\% of such respondents said that they were ``unlikely" or ``somewhat unlikely" to have identified the point without algorithmic assistance. We report the mean rank assigned to such points, where
a smaller rank indicates that the algorithm would prioritize
those points more for human inspection. We find that FlaSH consistently ranks these points near the top of its list (more so than other methods), indicating that FlaSH can usefully direct human attention to points that would have been missed otherwise. This is a result of FlaSH's emphasis on discovering trendline outliers, which our prototyping showed are difficult for humans to recognize in public health data streams.

 Overall, FlaSH's strength lies in leveraging specific features of public health data, a simple model class to meet deployment criteria, and an intuitive test statistic. The combination of these ideas is why FlaSH can scale to the data volumes required, perform well on traditional outlier detection metrics, especially compared to the best-performing deep learning models, and crucially, prioritize points for human review that would not have been discovered otherwise.

\section{Deployment and Lessons Learned}
Based on FlaSH's empirical performance and design, it has been deployed as part of Delphi's daily workflow since February 2023. It runs on selected streams, and an expert reviewer inspects the ranked, outlying points. To support this interaction, we added a dashboard where expert reviewers can visualize each of FlaSH's calculations before flagging them. As new types of irregularities arise, an analyst in the loop can modify FlaSH to detect those respective outliers.

\paragraph{Lessons Learned.} For outlier detection methods that produce actionable outputs, intuitive methods with informative outputs that explicitly navigate contextual nuances (like how FlaSH directly leverages the statistical properties of public health streams) innately enhance trust in method outputs that may also translate to performance gains. Additionally, method evaluations should consider expert-generated ground truth tasks that cover classification, because classification can be more straightforward for humans, and ranking, because thresholds for classification may vary.

\section{Related Works}
There are numerous outlier and anomaly detection methods \cite{blazquez2021review}, but recent advancements in the field focus on deep learning applications \cite{pang2021deep}. In our experiments, we find deep learning methods perform poorly on this task for various reasons. Accordingly, only a handful of real-time outlier streaming algorithms have been adapted for public health streaming data. 
Specifically, point outlier detection approaches for COVID-19 streams like \cite{jombart2021real,karadayi2020unsupervised,wang2021comparing,agarwal2022real} consider the \textit{nonstationarity} of the data streams but use simulations for evaluation or only consider a limited set of outlier categories. Hence, they are not fully applicable to our setting. Some source-specific COVID outlier detection methods \cite{dong2022johns} that operate on data streams before Delphi receives them do not have publicly-available methods, but the continued presence of irregularities in those streams that impacts Delphi stakeholders underscores the importance of FlaSH. 

\section{Conclusion}

This paper presents FlaSH, a practical framework for computationally assisted quality control in public health data streams. FlaSH creates a list of the most important outlying recent data points for domain experts to review by using simple models to explicitly account for the nuances of public health streaming data. In our experimental evaluation, which addressed some open design and evaluation challenges in unsupervised time series outlier detection, FlaSH scaled to the task requirements, outperformed other methods (including deep learning approaches) in traditional outlier detection metrics, and successfully prioritized points that would not have been discovered without algorithmic assistance. Our results demonstrate that effective, practical outlier detection systems require careful, user-informed design and sustained effort. These efforts will have considerable benefits for Delphi's stakeholders and, ultimately, for public health data users. 

\section*{Ethical Statement}
This research was conducted in accordance with the principles embodied in the Declaration of Helsinki and in accordance with local statutory requirements. All participants consented to the study and could exit the study at any time. Approval was granted from Carnegie Mellon Univeristy IRB number STUDY2022\_00000240. 

\section*{Acknowledgements}

We want to thank the whole Delphi team and members of the Carnegie Mellon community for their input to and support of this project. This work was supported by the Centers for Disease Control and Prevention of the U.S. Department of Health and Human Services (HHS) as part of a cooperative agreement funded solely by CDC/HHS under federal award identification number U01IP001121, ``Delphi Influenza Forecasting Center of Excellence". The contents are those of the authors and do not necessarily represent the official views of, nor an endorsement by, CDC/HHS or the U.S. Government. This material is based upon work supported by the National Science Foundation Graduate Research Fellowship under Grant No. DGE1745016 and DGE2140739. Any opinion, findings, and conclusions or recommendations expressed in this material are those of the authors and do not necessarily reflect the views of the National Science Foundation.

\bibliographystyle{acm}
\bibliography{bib}

\end{document}